\documentclass[10pt,twocolumn,letterpaper]{article}

\usepackage{iccv}
\usepackage{times}
\usepackage{epsfig}
\usepackage{graphicx}
\usepackage{amsmath}
\usepackage{amssymb}

\usepackage[breaklinks=true,bookmarks=false]{hyperref}

\iccvfinalcopy 

\def\iccvPaperID{****} 

\ificcvfinal\fi

\begin{document}

\title{Video Anomaly Detection By The Duality Of Normality-Granted Optical Flow}

\author{Hongyong Wang, Xinjian Zhang, Su Yang\\
Fudan University\\
{\tt\small \{hongyongwang19, zhangxj17, suyang\}@fudan.edu.cn}
\and
Weishan Zhang\\
China University of Petroleum\\
{\tt\small zhangws@upc.edu.cn}
}

\maketitle
\ificcvfinal\thispagestyle{empty}\fi

\begin{abstract}
Video anomaly detection is a challenging task because of diverse abnormal events. To this task, methods based on reconstruction and prediction are wildly used in recent works, which are built on the assumption that learning on normal data, anomalies cannot be reconstructed or predicated as good as normal patterns, namely the anomaly result with more errors. In this paper, we propose to discriminate anomalies from normal ones by the duality of normality-granted optical flow, which is conducive to predict normal frames but adverse to abnormal frames. The normality-granted optical flow is predicted from a single frame, to keep the motion knowledge focused on normal patterns. Meanwhile, We extend the appearance-motion correspondence scheme from frame reconstruction to prediction, which not only helps to learn the knowledge about object appearances and correlated motion, but also meets the fact that motion is the transformation between appearances. We also introduce a margin loss to enhance the learning of frame prediction. Experiments on standard benchmark datasets demonstrate the impressive performance of our approach.
\end{abstract}


\section{Introduction}

\begin{figure}[t]
\begin{center}
    \includegraphics[width=1\linewidth]{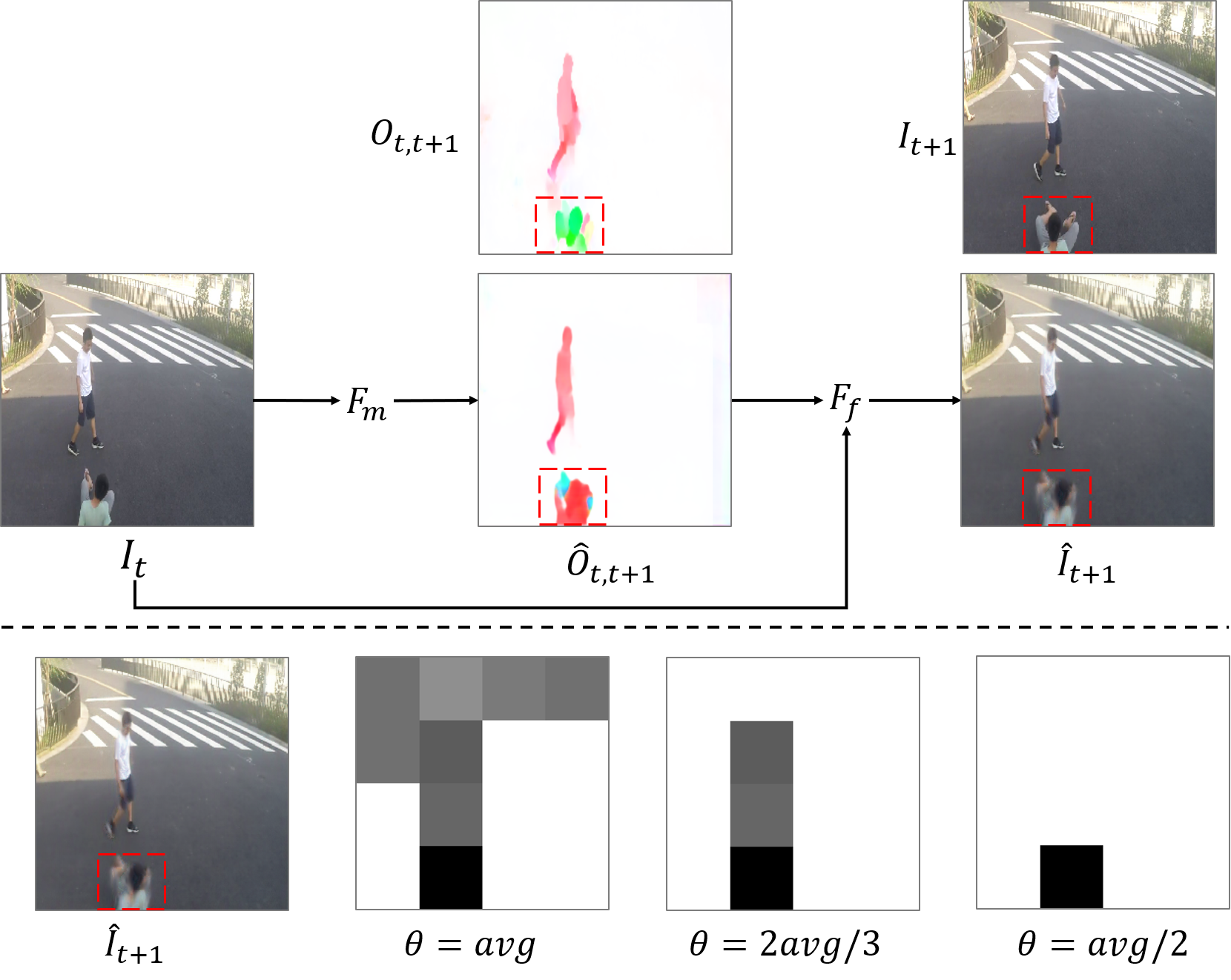}
\end{center}
   \caption{The effectiveness of duality of normality-granted optical flow in our model. As shown in the top part, the predicted optical flow of the abnormal behavior labeled by the red box is evidently different from the ground truth. But for the normal behavior of the near pedestrian, the predicted optical flow is relatively closer to the ground truth. This discrepancy will determine the optical flow guidance works for help or disturbance in the frame prediction, as the abnormal region in the predicted frame is more blurry. In the bottom part, we estimate the quality of the predicted frame by calculating the PSNR of patches with 64 \(\times\) 64 pixels. In PSNR masks, the darker patch suggests the higher confidence to anomalies, where we can see that the only one potential anomaly patch in the last mask meets the abnormal region in \({I}_{t+1}\).}
\label{fig: motivation}
\end{figure}

Along with the development of video surveillance nowadays, video anomaly detection (VAD) has drawn increasingly more attention. As an artificial intelligence technique, it will allow staff to work more efficiently because they don't have to pay attention to videos all the time anymore. A well-designed video anomaly detection system will inform staff only when there are potential abnormal events.

In this task, the anomaly is defined as unexpected events, related to scenes. Due to the rarely available labeled abnormal data, unsupervised learning is now playing a vital role. Similar to the common anomaly detection task, the training data only contains normal events, whereas both normal and abnormal events are included in the testing data. Unsupervised learning, here mainly including statistic \cite{M4}, reconstruction-based \cite{M10} and prediction-based \cite{R9} methods, will learn the normal patterns on the training dataset and score frames in the testing time. These methods work with the assumption that anomalies cannot be expressed well by normal patterns learned from the training data.

Now, reconstruction-based and prediction-based methods go in the spotlight, owing to the great success of deep learning. Reconstruction-based methods reconstruct hand-crafted features of a video clip, like \cite{M7}, or use Auto-Encoder to reconstruct a clip or single frame, like \cite{R12}. After works \cite{R9, FPN_2}, prediction-based methods are adopted by more and more works. Most of them are based on U-net \cite{Unet}, like \cite{R13, R11}. Especially, some of them combined reconstruction and prediction. \cite{R11} proposed a model that integrates reconstruction and prediction. \cite{unfair} proposed a model that can be applied for reconstruction or prediction.

As objects and their associated actions usually follow certain patterns, method \cite{R12} proposed to learn the appearance-motion correspondence by predicting optical flow from the current frame while reconstructing the frame, and achieved an improvement on the performance. However, there is the main weakness in their method. They force the current frame and optical flow to share the same latent space, which causes a temporal conflict because the optical flow is about the future but the frame is in the present. Another work \cite{R8} based on the optical flow and current frame also fell into such temporal conflict.

To overcome this drawback and better make use of the optical flow, we propose a novel prediction-based approach. We predict the next frame from the current one, guided by the normality-granted optical flow with the duality on normal and abnormal events, as shown in Fig.\ref{fig: motivation}. Our design brings three advantages. Firstly, we avert the temporal conflict in previous works, like above. Secondly, we conduct the appearance-motion learning scheme under frame prediction setting, which meets the nature of the optical flow, the pixel displacement between two adjacent frames. Thirdly and most importantly, the duality of normality-granted optical flow is leveraged to help distinguish normal and abnormal events in the frame prediction stage.

Our normality-granted optical flow concentrates on normal motion patterns, associated with normal objects. Learning from the single frame and only normal motion available in the training data, the model only can learn the translation from the spatial feature to normal motion mode. Hence, in the testing time, whatever the factual behaviors are normal or abnormal, the predicted optical flow will always belong to the normal mode.

This consistency will become duality in the frame prediction, when the optical flow is employed as the guidance, improving the ability of our model to discriminate between normal and abnormal frames. Concretely, for normal behaviors, the optical flow will offer helpful information, making the predicted frame close to the ground truth. But for abnormal behaviors, the optical flow becomes noise for the frame prediction, same for abnormal objects, which will result in more prediction errors in the predicted frame, with cases in Section.\ref{motion evaluation} and Section.\ref{frame evaluation}. Consequently, we can distinguish the abnormal frames from normal ones by prediction quality. Meanwhile, we make the other two improvements on frame prediction. To implement the appearance-motion correspondence scheme and align the object with its motion, we take a region-correspondence learning tactic, more robust to directly warp. To make the model focus on the motion of objects, we propose a margin loss, which will force the model to learn the difference between adjacent frames.

The main three contributions of our work as follows:
\begin{enumerate}
\item As far as we know, in the context of video anomaly detection, we are the first to leverage the normality-granted optical flow to guide the frame prediction, with whose duality the performance of our model is improved remarkably on distinguishing normal and abnormal frames.
\item To robustly make use of the optical flow, we introduce a region-correspondence learning tactic. To promote the model to learn the difference between adjacent frames, we propose a margin loss. These two introductions focusing on the transformation of frames will help the model to learn temporal features of events.
\item Experiments conducted on three main standard benchmarks demonstrate that we achieve the state-of-the-art on open-set video anomaly detection.
\end{enumerate}

\begin{figure*}
\begin{center}
\includegraphics[width=1\linewidth]{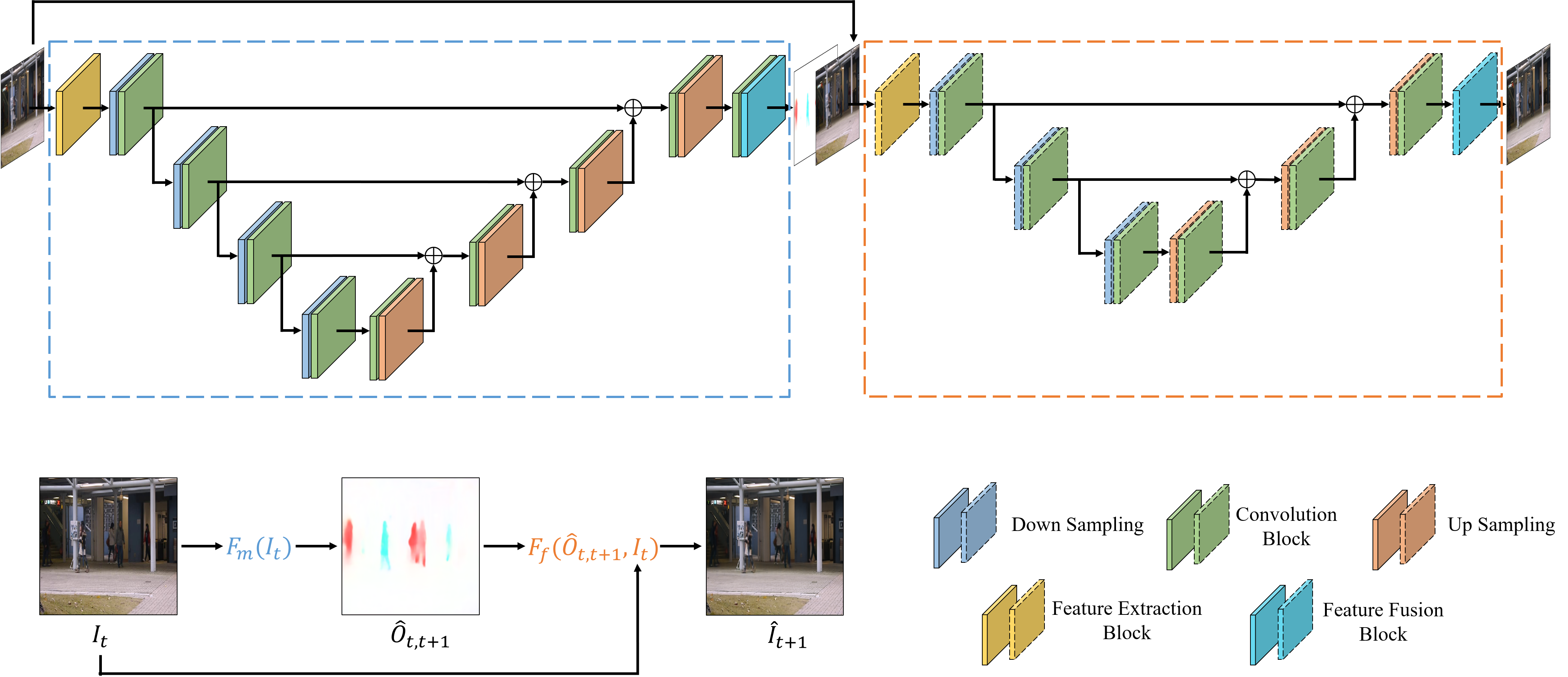}
\end{center}
   \caption{Overview of our network. Our model consists of two sub-networks, applied for motion and frame prediction respectively. Our motion prediction model \textcolor[RGB]{91,155,213}{\(F_{m}\)} marked by the \textcolor[RGB]{91,155,213}{blue} box is to predict the optical flow from the current frame. And guided by this optical flow, our frame prediction model \textcolor[RGB]{237,125,49}{\(F_{f}\)} marked by the \textcolor[RGB]{237,125,49}{orange} box will predict the next frame. Note that, our motion and frame prediction model are different in detail of every \textit{Block}, though they are the same name.}
\label{fig: overview}
\end{figure*}

\section{Related work}

{\bf Open-set VAD.} Open-set indicates unlimited potential anomalies, which is also organized as one-class classification like \cite{AddOneclass}. For this task, most methods are unsupervised including statistic \cite{M4,M5,M6}, reconstruction, and prediction based methods. It is worth pointing out that training only on normal data is often claimed as unsupervision like \cite{R10, M10, unfair} recently. Reconstruction methods learn the feature space by reconstructing normal features or frames. Early works based on reconstruction usually utilize hand-crafted features like HOG \cite{HOG} and sparse coding or dictionary learning to encode the normal patterns like \cite{M7,M8,M9}. Besides these, \cite{M1,M2} utilize low-level trajectory features and \cite{M3} exploits a Markov random filed to model normal patterns. In the context of deep learning, to limit the feature space around normal events, AutoEncoder is employed in many works \cite{M10,M11,R12}. By introducing RNN, \cite{M12,M13} use the ConvLSTM to model spatial-temporal features. After works \cite{R9,FPN_2}, prediction methods are adopted by more and more researchers. Prediction methods learn the normal patterns by 3D convolution \cite{3D}, U-net \cite{Unet}, or RNN. 3D convolution, applied in many video recognition tasks, focusing on both spatial and temporal patterns, has been proved its ability to learn motion information. U-net is a famous model structure for image segmentation \cite{image_seg_1} and image translation \cite{image_trans_1}. As frame prediction is a kind of image translation, many works recently take the modified U-net as their backbones. 


{\bf Close-set VAD.} Close-set indicates limited potential anomalies on objects, behaviors, or both. A typical close-set solution is object-centric which focuses on specific objects. Works \cite{object_centric_2, object_centric_4} propose to learn the normal poses of person. Works \cite{R6,object_centric_1,object_centric_5, AddCloseset2, AddCloseset_detail} try to find objects by pre-trained object detector, before analyzing events. Although \cite{AddCloseset_detail} utilize motion cues as complementary information, objects still are probably lost, especially the static. Whatever pose estimation, object detection, or semantic segmentation, they are only aware of limited objects, a subset of potential abnormal cases. But on the other hand, pre-detection gives the model an advantage with clearer information of objects, which benefits the model learning. 

Another common setting for the close-set problem is available labeled abnormal data with limited events in supervised methods. \cite{supervision_1} proposes a weakly labeled dataset for weakly supervised video anomaly detection. They take multiple instance learning (MIL) to train an anomaly ranking model on their video-level labeled dataset. \cite{supervision_2} mixes some abnormal data from the testing dataset with normal data to directly learn the difference between normal and abnormal events. Others like \cite{label_opt_1,label_opt_2} leverage and optimize results of other unsupervised or weakly supervised methods, which can be regarded as label denoise task. By the label, models are able to learn abstract knowledge of abnormal events handily, but their detection ability is limited to annotated specific events.

{\bf Optical flow.} Optical flow is a hand-crafted feature, wildly applied in various compute vision tasks, representing the motion information. With the remarkable performance of deep learning, some works have proposed deep learning-based methods, like FlowNet2 \cite{FlowNet2} and PWC-net \cite{PWC-net}, to estimate the optical flow and have achieved notable performance, with fast speed. It is common to see that optical flow is used in two-stream networks in video recognition task \cite{video_rec_2, video_rec_3}. As recognition and anomaly detection, to some extent, are similar in analyzing actions, optical flow also plays an important role in video anomaly detection. Work \cite{R9} adds an optical flow loss as the motion constraint during the training time. In \cite{R8,R12}, their models learn the motion by predicting the optical flow of the current frame. Optical flow guidance is also adopted by general video prediction tasks \cite{FramePredOpt1,FramePredOpt2,FramePredOpt3} to improve the performance, but we are the first to apply this idea to video anomaly detection task. There is a significant difference between our work and general video prediction works, that they try to predict all frames with high quality but we try to predict normal frames with high quality and abnormal frames with low quality.

\section{Approach}
\label{section: approach}

As shown in Fig.\ref{fig: overview}, our model is made up of two series sub-networks, the former one responsible for motion prediction and the latter for frame prediction. In the motion prediction, our model will predict the normality-granted optical flow from the current frame. This predicted optical flow concatenated with the current frame will be leveraged to predict the next frame in frame prediction. This structure conforms to the appearance-motion correspondence scheme, by from the appearance to motion and then to future appearance finally.

When an unexpected action occurs, our model is not able to predict its real corresponding optical flow and still generates the \textit{stereotypically normal} optical flow according to what it learns in the training time. In a word, for abnormal behaviors, the predicted optical flow will mismatch factual actions. These mismatches of optical flows will offer wrong guidance in the frame prediction, resulting in more prediction errors in the generated frame. In contrast, for normal frames, there will be fewer errors in the predicted frame. Ultimately, the anomaly score is computed according to prediction errors.

\subsection{Motion prediction}


Motion prediction is to predict the normality-granted optical flow. By converting the optical flow to RGB format, this subtask can be regarded as the image translation task, so we utilize U-net \cite{Unet}, with good performance at image translation \cite{R9}, as our basic architecture. The skip connection, an important component in U-net, will be instrumental in learning, since that optical flow and frame are not in the same domain. To better extract features from the frame, we replace the first two convolution layers with multi-scale convolution, named feature extraction block, consisting of four parallel convolution operations with different kernels. Meanwhile, we use the LeakyReLU instead of ReLU as our activation function, to achieve a better feature space. We also implement a deeper network to enhance the model to learn high-level semantic information.

To train our motion prediction model, we exploit the optical flow generated by PWC-net \cite{PWC-net} as the ground truth. PWC-net is a compact but effective CNN model for optical flow \cite{PWC-net}. However, there is a common shortcoming in deep learning-based optical flow estimation methods, that is, the inevitable noise in their results. In order to reduce the impact of noise, we employ the L1 distance as our loss function, for it is less sensitive to noise \cite{Fast_RCNN}.
\begin{equation}
L_{opt}(O_{t,t+1}, \hat{O}_{t,t+1}) = \| O_{t,t+1} - \hat{O}_{t,t+1} \|_{1}
\end{equation}
where \(O_{t,t+1}\) is the ground truth optical flow between frame \(I_{t}\) and \(I_{t+1}\), and \(\hat{O}_{t,t+1}\) is the optical flow predicted by our model from frame \(I_{t}\).

Our motion prediction can be formulated as
\begin{equation}
\hat{O}_{t,t+1} = F_{m}(I_{t})
\end{equation}
where \(F_{m}\) is our motion prediction model, the mapping function from current frame to optical flow.

\subsection{Frame prediction}


Frame prediction is to predict the next frame from the current one guided by the optical flow, generated from the previous stage. A simple way to implement the guidance is to warp the frame with optical flow, moving pixels by plus the displacements on \(x\) and \(y\) axes. This is suitable for clean and accurate optical flow, but as mentioned in the last section, our predicted optical flow suffers from the inevitable noise and is in RGB instead of coordinate format. Therefore, we take convolutions to conduct the guidance, to be more robust and powerful.

Concretely, we adopt a region-correspondence learning convolution tactic in the feature extraction block, by concatenating the frame and the optical flow map along channels. Thanks to the work manner of convolution kernels, these features are extracted from the corresponding regions of the frame and the optical flow, responding to the relation between objects and their motion. By this tactic, our model can learn features from the frame and optical flow at the same time and then fuse them. 

As the U-net is wildly applied to predict frame in video anomaly detection, our frame prediction model is also based on the U-net. Same as the purpose in motion prediction, we exploit the LeakyReLU as the activation function. To offset the cost of our deeper motion prediction model, we have cut down layers of our frame prediction model, which is empirically still competent to prediction task.

Like many other works on image translation, we employ L2 distance as one of our loss functions. Besides, as the motion is represented by the difference between frames, to allow our model to be focused on behaviors, we introduce other margin loss, modified from the Triplet Loss \cite{Triplet}, to force the model to learn the transformation between adjacent frames. In other words, L2 loss is to make the prediction closer to the ground truth, while margin loss impels prediction to be farther from the current frame.
\begin{equation}
L_{dense}(I_{t+1}, \hat{I}_{t+1}) = \| I_{t+1} - \hat{I}_{t+1} \|_{2}
\end{equation}
\begin{eqnarray}
L_{margin}(I_{t+1}, \hat{I}_{t+1}, I_{t}) = Triplet(I_{t+1},\hat{I}_{t+1},{I}_{t}) \nonumber \\
= max(0, \| I_{t+1} - \hat{I}_{t+1} \|_{2} - \| I_{t} - \hat{I}_{t+1} \|_{2} + \alpha ) \label{margin loss}
\end{eqnarray}
where \(I_{t}\) and \(I_{t+1}\) is the ground truth frame at time \(t\) and \(t+1\). \(\hat{I}_{t+1}\) is the predicted frame at time \(t+1\), and \(\alpha\) is the expected distance between the predicted and current frame. The final loss for frame prediction is shown below:
\begin{equation}
L_{frame} = L_{dense} + \lambda{L}_{margin}
\end{equation}

Our frame prediction sub-network can be formulated as
\begin{equation}
\hat{I}_{t+1} = F_{f}(\hat{O}_{t,t+1}, I_{t})
\end{equation}
where \(F_{f}\) is our frame prediction model.

At the end, our whole network can be formulated as
\begin{gather}
\hat{I}_{t+1} = F(I_{t}) \\
F(I_{t}) = F_{f}(F_{m}(I_{t}), I_{t})
\end{gather}
where \(F\) is our whole network, the mapping function from the current frame \(I_{t}\) to the next one \(\hat{I}_{t+1}\).

\subsection{Anomaly detection}

Anomaly detection is to score a frame on account of its possibility to be abnormal. As mentioned, for abnormal frames, there will be more prediction errors, which will impact the quality of the predicted frame, compared with the ground truth. Therefore, it is intuitive to calculate the anomaly score by measuring image quality. The same as \cite{R9}, we use the Peak Signal to Noise Ratio (PSNR) for image quality assessment.
\begin{equation}
PSNR(I, \hat{I}) = 10\log_{10}\begin{bmatrix}\frac{MAX^{2}_{I}}{MSE(I,\hat{I})}\end{bmatrix}
\end{equation}
\begin{equation}
MSE(I,\hat{I}) = \frac{1}{M}\frac{1}{N}\sum_{i=1}^{M}\sum_{j=1}^{N}(I_{i,j}-\hat{I}_{i,j})^2
\end{equation}
where \(I\) is the ground truth, and \(\hat{I}\) is the predicted frame. \(M, N\) is the height and width of frames separately. \(MAX_{I}\) is the maximum pixel value of the ground truth frame.

A higher PSNR value indicates a higher quality of the frame, in other words, more likely to be normal. Then, to rescale the score in \([0, 1]\), we normalize the PSNR value in each video and denote the result as \(NS\), the normality score.
\begin{equation}
NS_{t} = \frac{P_{t} - P_{min}}{P_{max} - P_{min}}
\end{equation}
where, for frame at time \(t\), \(NS_{t}\) is its normalized PSNR value, and \(P_{t}\) is its original PSNR value. The \(P_{max}\) and \(P_{min}\) is the max and min PSNR value in the current video.

The final step is to convert \(NS\) to the anomaly score. 
\begin{equation}
S_{t} = 1 - NS_{t}
\end{equation}
where \(S_{t}\) is the anomaly score of the frame at time \(t\).

As the prediction result starts from the second frame, we directly give the first frame a score as same as the second frame.

\section{Experiments}

\subsection{Dataset}

{\bf UCSD Ped2.} The UCSD Ped2 dataset \cite{Ped2} contains 16 training videos and 12 testing videos with 12 abnormal events. Anomalies are vehicles like bicycles, cars, and wheelchairs, crossing the pedestrian areas.

{\bf CUHK Avenue.} The CUHK Avenue dataset \cite{M8}, a very popular and challenging dataset for video anomaly detection, contains 16 training and 21 testing video clips. All videos are captured in a fixed scene but with some hard abnormal events. 

{\bf ShanghaiTech.} The ShanghaiTech dataset \cite{R7} is the largest and challenging dataset. There are 330 training videos and 107 testing videos. Different from the above datasets, it consists of 13 scenes with 130 abnormal events, making the task more difficult. Meanwhile, anomalies caused by sudden motion are included in this dataset, making this dataset closer to real scenarios. These two characteristics make this dataset more challenging.

\begin{table}[t]
\begin{center}
\begin{tabular}{|c|c|c|c|}
\hline
Method                & UCSD Ped2      & Avenue        & ShanghaiTech  \\ \hline\hline
MPPCA \cite{M5}        & 69.3           & -             & -             \\ \hline
MDT \cite{M6}          & 82.9           & -             & -             \\ \hline
ConvAE \cite{M10}      & 85.0           & 80.0          & 60.9          \\ \hline
AMDN \cite{R4}         & 90.8           & -             & -             \\ \hline
Unmasking \cite{R5}    & 82.2           & 80.6          & -             \\ \hline
StackRNN \cite{R7}     & 92.2           & 81.7          & 68.0          \\ \hline
AbnormalGAN \cite{R8}  & 93.5           & -             & -             \\ \hline
OWC \cite{AddRank1}    & 94.0           & -             & -             \\ \hline
FFPN \cite{R9}         & 95.4           & 85.1          & 72.8          \\ \hline
MemAE \cite{R10}       & 94.1           & 83.3          & 71.2          \\ \hline
AMC \cite{R12}         & 96.2           & 86.9          & -             \\ \hline
AnoPCN \cite{R13}      & \textbf{96.8}  & 86.2          & 73.6          \\ \hline
AnomalyNet \cite{R14}  & 94.9           & 86.1          & -             \\ \hline
Integrating \cite{R11} & 96.3           & 85.1          & 73.0          \\ \hline
CDDA \cite{R15}        & 96.5           & 86.0          & 73.3          \\ \hline
Ours                  & 94.2           & \textbf{88.4} & \textbf{75.3} \\ \hline
\end{tabular}
\end{center}
\caption{Comparison with the state-of-the-art on frame-level performance on UCSD Ped2, CUHK Avenue, and ShanghaiTech dataset. The best performance is in bold.}
\label{table: rank}
\end{table}

\subsection{Training}

Every frame from different datasets is resized to \(256 \times 256\) and normalized to \([0, 1]\). For optical flow, we use the PWC-net \cite {PWC-net} pre-trained on FlyingChairs dataset \cite{FlyingChairs} as the ground truth estimator. In the pre-processing, the optical flow generated by PWC-net will be converted to the RGB format, and then resized to \(256 \times 256\) and normalized to [0, 1]. A typical seeting for \(\alpha\) and \(\lambda\) in the loss function is \(0.2\) and \(0.004\). Although our two sub-networks are different in the task, we basically adopt the same training settings for both stages, due to that we focus on the methodology rather than hyperparameters. We apply the Adam \cite{Adam} optimizer with \(\beta_{1}=0.9 \) and \(\beta_{2}=0.999\), learning rate as \(2 \times 10^{-4} \), and batch size as 16 for all datasets on two stages, except the UCSD Ped2 in frame prediction, whose learning rate is \(2 \times 10^{-5} \). As our network, a two-stage model, consists of two series sub-networks, we first train the motion prediction sub-network 10, 20, 20 epochs for ShanghaiTech, CHUK Avenue, and UCSD Ped2 respectively, and then select the result to help the frame prediction. For frame prediction, we train sub-network 10 epochs on all datasets. All experiments are conducted with Nvidia GTX 1080Ti.

\subsection{Evaluation}

According to the convention, we leverage the Area Under Curve (AUC) of the Receiver Operation Characteristics (ROC), namely AUC@ROC, as the evaluation metric. We evaluate our model on frame-level video anomaly detection. For the frame annotated at pixel-level and without the frame-level label, we mark it as abnormal if any pixel of it is annotated as abnormal.


\subsection{Result}

We compare our method with the state-of-the-art, shown in Table.\ref{table: rank}. To be fair, we follow \cite{R9,R12}, which normalize the final score to \([0, 1]\) in each video and then concatenated to calculate the AUC score, without any else post-processing. As the table shows, we achieve the AUC at 88.4 on CHUK Avenue and 75.3 on ShanghaiTech, being the best performance on these two datasets, which demonstrates the effectiveness of our method.

For UCSD Ped2, a related reason for our not perfect result is that the PWC-net, our ground truth generator for optical flow, doesn't work well on this dataset sometimes, which can be found in the middle column of the last two rows in Fig.\ref{fig: ped2_motion_noise}. There is much noise in the background and part of the generated optical flow for pedestrians is mixed up. Though we don't achieve the state-of-the-art, we have got a great performance, our score more than \(90\), with the ceiling in \(100\), which is enough for application, where moreover we achieve the best on the other two datasets.


\subsection{Motion evaluation}

\label{motion evaluation}

\begin{figure}[t]
\begin{center}
    \includegraphics[width=1\linewidth]{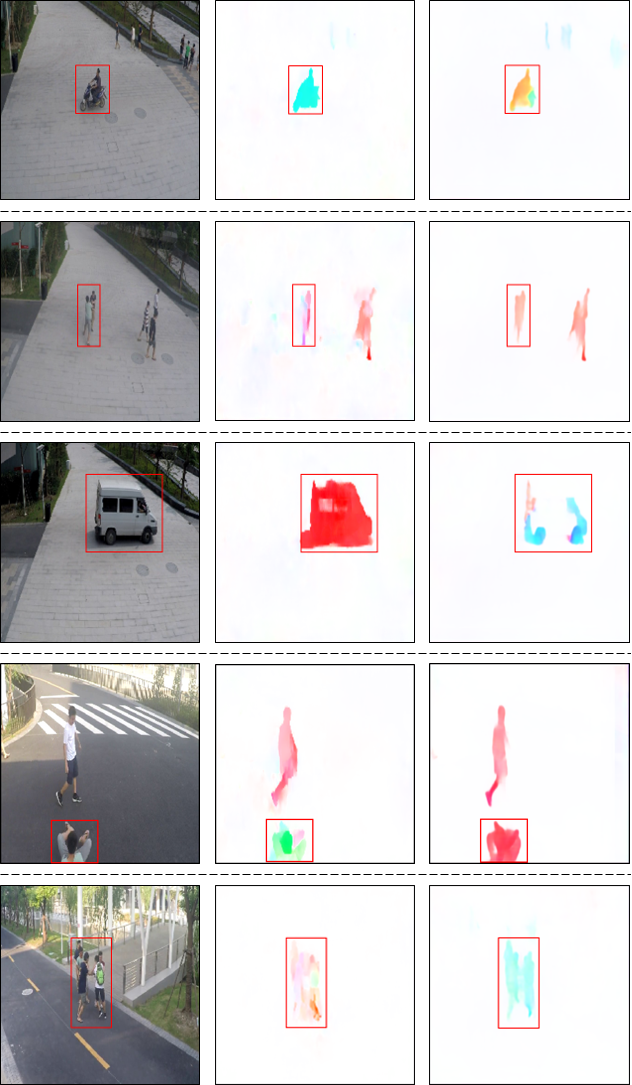}
\end{center}
   \caption{Illustration of the motion prediction for abnormal frames on the ShanghaiTech dataset. Each row consists of three columns. The left column is the input frame, noted as \(I_{t}\) in our paper. The middle column is the ground truth of the optical flow between \(I_{t}\) and \(I_{t+1}\), noted as \(O_{t,t+1}\). And the right column is the predicted optical flow from frame \(I_{t}\), noted as \(\hat{O}_{t,t+1}\). Abnormal regions are labeled by red rectangles. We can see that there is a distinct difference between our prediction and ground truth. This conflict will aid our model in finding anomalies. Best viewed in color.}
\label{fig: opt_comparison_abnormal}
\end{figure}

\begin{figure}[t]
\begin{center}
    \includegraphics[width=1\linewidth]{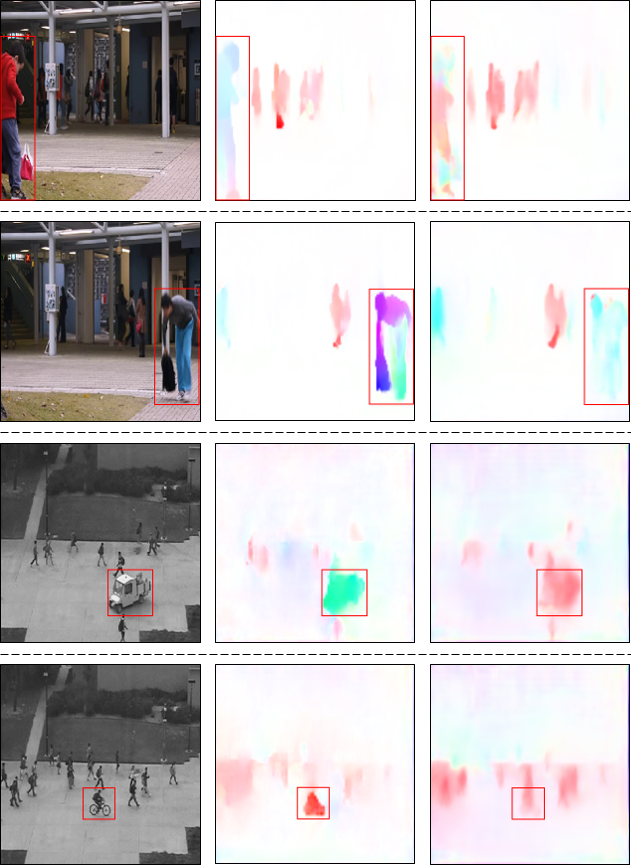}
\end{center}
   \caption{Illustration of the motion prediction for abnormal frames on Avenue and Ped2 dataset. Each column has been organized as same as Fig.\ref{fig: opt_comparison_abnormal}. The top two rows from the CUHK Avenue dataset and the last two rows from the UCSD Ped2 dataset. Best viewed in color.}
\label{fig: opt_comparison_abnormal_ave_and_ped}
\label{fig: ped2_motion_noise}
\end{figure}

We select some examples to show the ability of our model to distinguish anomaly from normality at the motion level, for ShanghaiTech dataset shown in Fig.\ref{fig: opt_comparison_abnormal}, and for Avenue and Ped2 datasets shown in Fig.\ref{fig: opt_comparison_abnormal_ave_and_ped}. The number of samples selected from different datasets is concerned with the scale and the anomaly diversity of the dataset. The ground truth for optical flow is generated by the PWC-net. Video results can be found in the supplementary.

There are five typical events from the ShanghaiTech dataset, shown in Fig.\ref{fig: opt_comparison_abnormal}. The abnormal event in the first row is a person riding a motorcycle, annotated by the red rectangle. Because our model has never seen the motorcycle in the training time, it is hard for it to predict the correct optical flow. In the second row, two students are brawling. For this abnormal behavior, our model will predict the optical flow following the knowledge learned in the training time. Therefore, the prediction will be different from the ground truth, because motion patterns of this abnormal behavior are not included in the training data. Meanwhile, for near pedestrians, our prediction result is better at the edge of the optical flow. The car is the abnormal object in the third row, and the predicted optical flow is corrupted. In the fourth row, a student is tumbling over backward, but our model thinks that he is moving to the right. This contradiction is depicted by the optical flow and will help our network to detect abnormal behaviors. The last row is another brawl event but with more students. From the figure, we can find that our model predicts a mismatched optical flow to the group of students, demonstrating the ability of our model to handle group events.

\begin{figure}[t]
\begin{center}
    \includegraphics[width=1\linewidth]{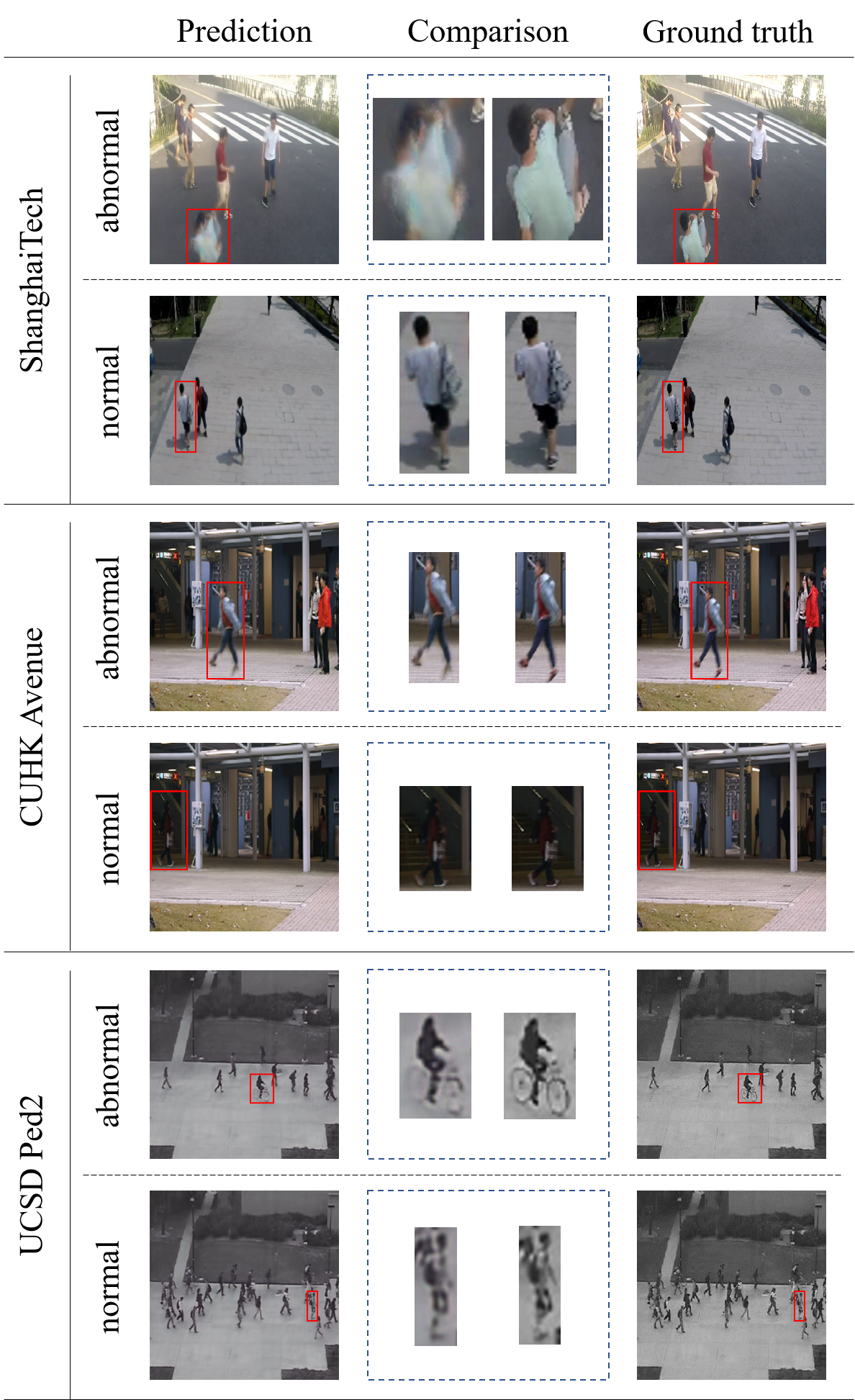}
\end{center}
   \caption{Illustration of the frame prediction. For every evaluation dataset, we have selected one abnormal event and one normal event. And for every frame, the comparison region is noted by the red box. To be clear, we resize the marked areas at the middle column, which, named comparison column, always holds two images, the left one from the prediction and the right one from the ground truth. As shown, there are more prediction errors with abnormal events.}
\label{fig: frame_comparison}
\end{figure}

For CUHK Avenue dataset, two typical abnormal events are shown in the top two rows in Fig.\ref{fig: opt_comparison_abnormal_ave_and_ped}. In the first row, a child walking on the lawn, labeled as abnormal, is distinguished by the large difference between the two optical flow maps. In the second row is a man bending to pick up his backpack, whose action is depicted by the ground truth optical flow. But our model predicts that he is just moving to the left. For abnormal behaviors, our model will give it an optical flow in normal mode, conflicted with the actual actions, by which the anomaly will be distinguished.

For UCSD Ped2 dataset, we select two typical abnormal samples shown in the last two rows in Fig.\ref{fig: opt_comparison_abnormal_ave_and_ped}. In the third row, the anomaly is the moving car. From the shown optical flows, we can observe that our model predicts an incorrect motion direction to the car. Due to that our model has never learned the patterns of the abnormal object, the car, our model has mistaken its motion trend, by which this anomaly will be distinguished. The abnormal event in the last row is people riding a bike. Although its direction is the same as surrounding normal pedestrians, it moves faster, represented by the darker red color. The difference in luminance points out the potential anomaly.

On the other hand, our model can predict the motion of normal patterns well. In Fig.\ref{fig: opt_comparison_abnormal} and  Fig.\ref{fig: opt_comparison_abnormal_ave_and_ped}, there are normal cases that have been predicted well, similar to the ground truth. What our model learns are typical motion patterns, so, for normal behaviors, the predicted optical flow is not exactly the same as the ground truth sometimes. Even so, the discrepancy in the optical flow of abnormal events is more manifest in the same dataset, which will lead to more errors in frame prediction.

\subsection{Frame evaluation}

\label{frame evaluation}

To directly show the difference between predictions of abnormal and normal events, we have selected one abnormal event and one normal event from every evaluation dataset, shown in Fig.\ref{fig: frame_comparison}. Video results can be found in the supplementary.

For ShanghaiTech, the abnormal event is a student tumbling over backward, of which the prediction is distorted, which will make the quality of the predicted image lower. Compared with the ground truth through the PSNR function, there will be a small evaluation value, suggesting that this frame is more likely to be abnormal. Meanwhile, our model can predict the normal event better, like the pedestrian and the texture of his bag, shown in the normal row.

For CUHK Avenue, the abnormal event is the jumping kid. We can see that one of the predicted shoes is in the opposite direction and the other one almost disappears, which both indicate that this behavior is not able to be constructed by normal patterns well. In the normal row, the shoes and the bag are predicted better.

For USCD Ped2, though we didn't make a new state-of-the-art, we have got a good result. Here we show a typical anomaly in this dataset, that riding a bicycle. From the comparison, we can see that parts of the abnormal object almost vanish. In other words, those parts cannot be constructed appropriately by normal patterns. But in the normal row, for the pedestrian, the texture of shorts is predicted better, although it is not exactly the same as the ground truth.

Undeniably, there are also errors in the normal prediction, holding back us from being perfect. Although our method brings more errors into predicted abnormal frames, it can't guarantee that no errors in the normal ones. But, most of the time, our model will predict worse results for abnormal frames. Worse results will lead to lower scores under PSNR function, namely, higher confidences to be abnormal.

\subsection{Ablation study}

\begin{table}[h]
\begin{center}
\begin{tabular}{|c|c|c|c|c|}
\hline
             & Exp1  & Exp2  & Exp3  & Exp4 \\ \hline
margin loss  &  -     & \checkmark     & -     & \checkmark    \\ \hline
optical flow & -     & -     & \checkmark     & \checkmark    \\ \hline
AUC     & 84.55 & 85.56 & 87.45 & 88.41   \\ \hline
\end{tabular}
\end{center}
\caption{Evaluation of two introduced ingredients in our method on CHUK Avenue dataset. The result of different combinations of two factors is reported. The \checkmark in every Exp column indicates that row ingredient being equipped.}
\label{table: ablation}
\end{table}

To show the effectiveness of two introduced ingredients, the margin loss and optical flow guidance, we conduct ablation experiments and report in Table.\ref{table: ablation}, employing the AUC as the performance metric. As motion information is the transformation in consecutive frames, we set the margin loss to force the model to learn the difference of two adjacent frames, which will separate the prediction from the input frame. From the table, we can see that it improves the performance by \(1\%\) or so. 

As for the optical flow guidance, it is a key component in our model. Firstly, by utilizing the optical flow, we can predict the motion from a single frame, learning the common motion patterns from the appearances. Secondly, the duality of normality-granted optical flow will contribute to anomaly detection. Lastly, optical flow is an explicit motion condition to promote the frame prediction. As shown in the table, compared to the valina model in Exp.1, our optical flow guidance gives about a \(3\%\) improvement on performance, demonstrating the effectiveness.

\subsection{Running time}

\begin{table}[h]
\begin{center}
\begin{tabular}{|c|c|c|c|}
\hline
           & UCSD Ped2 & Avenue & ShanghaiTech \\ \hline
prediction & 76 fps    & 77 fps & 77 fps       \\ \hline
detection  & 47 fps    & 47 fps & 47 fps       \\ \hline
\end{tabular}
\end{center}
\caption{Inference speed of our model. We report the fps of our model on three evaluation datasets separately. The prediction row is the model inference speed, from the input to the predicted frame. After prediction, there is a step to calculate the anomaly score. The total cost of predicting and scoring is recorded as detection.}
\label{table: fps}
\end{table}

We report the running time of our method for prediction and detection in Table.\ref{table: fps}. As mentioned, there are two steps in our method to detect anomalies, that we predict a frame firstly and then give it an anomaly score. In the first step, denoted as the prediction in the table, the average running time is about 77 fps. And that for detection, covering two steps, is about 47 fps, achieving the real-time speed. Experiments are conducted on a single one NVIDIA GeForce 1080Ti GPU with batch size set to 1.

\section{Conclusion}

We have proposed a novel prediction-based method for open-set video anomaly detection. We employ the duality of normality-granted optical flow to discriminate between normal and abnormal frames. Meanwhile, our method extends the appearance-motion correspondence learning scheme from reconstruction to prediction and makes good use of the nature of the optical flow, which both are implemented by the region-correspondence tactic with the help of the convolution mechanism. We also introduce a margin loss to improve the frame prediction by forcing the model to learn the difference between the current frame and the next. Studies shown in this paper have demonstrated the effectiveness of our design and experiments on three standard benchmark datasets prove that our model achieves the state-of-the-art.

{\small
\bibliographystyle{ieee_fullname}
\bibliography{final}

\begin{thebibliography}{10}\itemsep=-1pt

\bibitem{M4}
A. {Adam}, E. {Rivlin}, I. {Shimshoni}, and D. {Reinitz}.
\newblock Robust real-time unusual event detection using multiple
  fixed-location monitors.
\newblock {\em IEEE Transactions on Pattern Analysis and Machine Intelligence},
  30(3):555--560, 2008.

\bibitem{image_seg_1}
V. {Badrinarayanan}, A. {Kendall}, and R. {Cipolla}.
\newblock Segnet: A deep convolutional encoder-decoder architecture for image
  segmentation.
\newblock {\em IEEE Transactions on Pattern Analysis and Machine Intelligence},
  39(12):2481--2495, 2017.

\bibitem{R15}
Yunpeng Chang, Zhigang Tu, Wei Xie, and Junsong Yuan.
\newblock Clustering driven deep autoencoder for video anomaly detection.
\newblock {\em 2020 European Conference on Computer Vision (ECCV)}, 2020.

\bibitem{M12}
Y.~S. Chong and Yong~Haur Tay.
\newblock Abnormal event detection in videos using spatiotemporal autoencoder.
\newblock {\em ArXiv}, abs/1701.01546, 2017.

\bibitem{M7}
Yang Cong, Junsong Yuan, and J. Liu.
\newblock Sparse reconstruction cost for abnormal event detection.
\newblock {\em CVPR 2011}, pages 3449--3456, 2011.

\bibitem{HOG}
N. {Dalal} and B. {Triggs}.
\newblock Histograms of oriented gradients for human detection.
\newblock In {\em 2005 IEEE Computer Society Conference on Computer Vision and
  Pattern Recognition (CVPR'05)}, volume~1, pages 886--893 vol. 1, 2005.

\bibitem{AddCloseset2}
Keval Doshi and Yasin Yilmaz.
\newblock Continual learning for anomaly detection in surveillance videos.
\newblock In {\em Proceedings of the IEEE/CVF Conference on Computer Vision and
  Pattern Recognition (CVPR) Workshops}, June 2020.

\bibitem{FlyingChairs}
A. Dosovitskiy, P. Fischer, E. Ilg, P. H{\"a}usser, C. Haz{\i}rba{\c{s}}, V.
  Golkov, P. v.d. Smagt, D. Cremers, and T. Brox.
\newblock Flownet: Learning optical flow with convolutional networks.
\newblock In {\em IEEE International Conference on Computer Vision (ICCV)},
  2015.

\bibitem{video_rec_3}
Christoph Feichtenhofer, Haoqi Fan, Jitendra Malik, and Kaiming He.
\newblock Slowfast networks for video recognition.
\newblock In {\em Proceedings of the IEEE/CVF International Conference on
  Computer Vision (ICCV)}, October 2019.

\bibitem{FramePredOpt2}
H. Gao, Huazhe Xu, Qi-Zhi Cai, R. Wang, F. Yu, and T. Darrell.
\newblock Disentangling propagation and generation for video prediction.
\newblock {\em 2019 IEEE/CVF International Conference on Computer Vision
  (ICCV)}, pages 9005--9014, 2019.

\bibitem{Fast_RCNN}
Ross Girshick.
\newblock Fast r-cnn.
\newblock In {\em Proceedings of the IEEE International Conference on Computer
  Vision (ICCV)}, December 2015.

\bibitem{R10}
Dong Gong, L. Liu, Vuong Le, Budhaditya Saha, M. Mansour, S. Venkatesh, and
  A.~V.~D. Hengel.
\newblock Memorizing normality to detect anomaly: Memory-augmented deep
  autoencoder for unsupervised anomaly detection.
\newblock {\em 2019 IEEE/CVF International Conference on Computer Vision
  (ICCV)}, pages 1705--1714, 2019.

\bibitem{M10}
M. Hasan, Jonghyun Choi, J. Neumann, A. Roy-Chowdhury, and L. Davis.
\newblock Learning temporal regularity in video sequences.
\newblock {\em 2016 IEEE Conference on Computer Vision and Pattern Recognition
  (CVPR)}, pages 733--742, 2016.

\bibitem{R6}
Ryota Hinami, T. Mei, and Shin'ichi Satoh.
\newblock Joint detection and recounting of abnormal events by learning deep
  generic knowledge.
\newblock {\em 2017 IEEE International Conference on Computer Vision (ICCV)},
  pages 3639--3647, 2017.

\bibitem{FlowNet2}
Eddy Ilg, Nikolaus Mayer, Tonmoy Saikia, Margret Keuper, Alexey Dosovitskiy,
  and Thomas Brox.
\newblock Flownet 2.0: Evolution of optical flow estimation with deep networks.
\newblock {\em 2017 IEEE Conference on Computer Vision and Pattern Recognition
  (CVPR)}, Jul 2017.

\bibitem{object_centric_1}
Radu~Tudor Ionescu, Fahad~Shahbaz Khan, Mariana-Iuliana Georgescu, and Ling
  Shao.
\newblock Object-centric auto-encoders and dummy anomalies for abnormal event
  detection in video.
\newblock In {\em Proceedings of the IEEE/CVF Conference on Computer Vision and
  Pattern Recognition (CVPR)}, June 2019.

\bibitem{R5}
Radu~Tudor Ionescu, Sorina Smeureanu, B. Alexe, and M. Popescu.
\newblock Unmasking the abnormal events in video.
\newblock {\em 2017 IEEE International Conference on Computer Vision (ICCV)},
  pages 2914--2922, 2017.

\bibitem{3D}
S. {Ji}, W. {Xu}, M. {Yang}, and K. {Yu}.
\newblock 3d convolutional neural networks for human action recognition.
\newblock {\em IEEE Transactions on Pattern Analysis and Machine Intelligence},
  35(1):221--231, 2013.

\bibitem{M5}
J. Kim and K. Grauman.
\newblock Observe locally, infer globally: A space-time mrf for detecting
  abnormal activities with incremental updates.
\newblock {\em 2009 IEEE Conference on Computer Vision and Pattern
  Recognition}, pages 2921--2928, 2009.

\bibitem{Adam}
Diederik Kingma and Jimmy Ba.
\newblock Adam: A method for stochastic optimization.
\newblock {\em International Conference on Learning Representations}, 12 2014.

\bibitem{Ped2}
W. {Li}, V. {Mahadevan}, and N. {Vasconcelos}.
\newblock Anomaly detection and localization in crowded scenes.
\newblock {\em IEEE Transactions on Pattern Analysis and Machine Intelligence},
  36(1):18--32, 2014.

\bibitem{FramePredOpt1}
Xiaodan Liang, L. Lee, Wei Dai, and E. Xing.
\newblock Dual motion gan for future-flow embedded video prediction.
\newblock {\em 2017 IEEE International Conference on Computer Vision (ICCV)},
  pages 1762--1770, 2017.

\bibitem{AddRank1}
Hanhe Lin, Jeremiah~D. Deng, Brendon~J. Woodford, and Ahmad Shahi.
\newblock Online weighted clustering for real-time abnormal event detection in
  video surveillance.
\newblock In {\em Proceedings of the 24th ACM International Conference on
  Multimedia}, MM '16, page 536–540, New York, NY, USA, 2016. Association for
  Computing Machinery.

\bibitem{supervision_2}
Wen Liu, Weixin Luo, Zhengxin Li, Peilin Zhao, and Shenghua Gao.
\newblock Margin learning embedded prediction for video anomaly detection with
  a few anomalies.
\newblock In {\em Proceedings of the Twenty-Eighth International Joint
  Conference on Artificial Intelligence, {IJCAI-19}}, pages 3023--3030.
  International Joint Conferences on Artificial Intelligence Organization, 7
  2019.

\bibitem{R9}
W. Liu, Weixin Luo, Dongze Lian, and Shenghua Gao.
\newblock Future frame prediction for anomaly detection - a new baseline.
\newblock {\em 2018 IEEE/CVF Conference on Computer Vision and Pattern
  Recognition}, pages 6536--6545, 2018.

\bibitem{M8}
Cewu Lu, J. Shi, and J. Jia.
\newblock Abnormal event detection at 150 fps in matlab.
\newblock {\em 2013 IEEE International Conference on Computer Vision}, pages
  2720--2727, 2013.

\bibitem{M13}
Weixin Luo, W. Liu, and Shenghua Gao.
\newblock Remembering history with convolutional lstm for anomaly detection.
\newblock {\em 2017 IEEE International Conference on Multimedia and Expo
  (ICME)}, pages 439--444, 2017.

\bibitem{R7}
Weixin Luo, W. Liu, and Shenghua Gao.
\newblock A revisit of sparse coding based anomaly detection in stacked rnn
  framework.
\newblock {\em 2017 IEEE International Conference on Computer Vision (ICCV)},
  pages 341--349, 2017.

\bibitem{M6}
V. Mahadevan, Weixin Li, Viral Bhalodia, and N. Vasconcelos.
\newblock Anomaly detection in crowded scenes.
\newblock {\em 2010 IEEE Computer Society Conference on Computer Vision and
  Pattern Recognition}, pages 1975--1981, 2010.

\bibitem{object_centric_4}
Amir Markovitz, Gilad Sharir, Itamar Friedman, Lihi Zelnik-Manor, and Shai
  Avidan.
\newblock Graph embedded pose clustering for anomaly detection.
\newblock In {\em IEEE/CVF Conference on Computer Vision and Pattern
  Recognition (CVPR)}, June 2020.

\bibitem{object_centric_2}
Romero Morais, Vuong Le, Truyen Tran, Budhaditya Saha, Moussa Mansour, and
  Svetha Venkatesh.
\newblock Learning regularity in skeleton trajectories for anomaly detection in
  videos.
\newblock In {\em Proceedings of the IEEE/CVF Conference on Computer Vision and
  Pattern Recognition (CVPR)}, June 2019.

\bibitem{R12}
T. Nguyen and J. Meunier.
\newblock Anomaly detection in video sequence with appearance-motion
  correspondence.
\newblock {\em 2019 IEEE/CVF International Conference on Computer Vision
  (ICCV)}, pages 1273--1283, 2019.

\bibitem{label_opt_1}
Guansong Pang, Cheng Yan, Chunhua Shen, Anton van~den Hengel, and Xiao Bai.
\newblock Self-trained deep ordinal regression for end-to-end video anomaly
  detection.
\newblock In {\em IEEE/CVF Conference on Computer Vision and Pattern
  Recognition (CVPR)}, June 2020.

\bibitem{unfair}
Hyunjong Park, Jongyoun Noh, and Bumsub Ham.
\newblock Learning memory-guided normality for anomaly detection.
\newblock In {\em IEEE/CVF Conference on Computer Vision and Pattern
  Recognition (CVPR)}, June 2020.

\bibitem{R8}
Mahdyar Ravanbakhsh, Moin Nabi, E. Sangineto, L. Marcenaro, C. Regazzoni, and
  N. Sebe.
\newblock Abnormal event detection in videos using generative adversarial nets.
\newblock {\em 2017 IEEE International Conference on Image Processing (ICIP)},
  pages 1577--1581, 2017.

\bibitem{Unet}
Olaf Ronneberger, Philipp Fischer, and Thomas Brox.
\newblock U-net: Convolutional networks for biomedical image segmentation.
\newblock {\em Medical Image Computing and Computer-Assisted Intervention –
  MICCAI 2015}, page 234–241, 2015.

\bibitem{AddOneclass}
Mohammad Sabokrou, Mohammad Khalooei, Mahmood Fathy, and Ehsan Adeli.
\newblock Adversarially learned one-class classifier for novelty detection.
\newblock In {\em Proceedings of the IEEE Conference on Computer Vision and
  Pattern Recognition (CVPR)}, June 2018.

\bibitem{Triplet}
Florian Schroff, Dmitry Kalenichenko, and James Philbin.
\newblock Facenet: A unified embedding for face recognition and clustering.
\newblock {\em 2015 IEEE Conference on Computer Vision and Pattern Recognition
  (CVPR)}, Jun 2015.

\bibitem{supervision_1}
Waqas Sultani, Chen Chen, and Mubarak Shah.
\newblock Real-world anomaly detection in surveillance videos.
\newblock In {\em Proceedings of the IEEE Conference on Computer Vision and
  Pattern Recognition (CVPR)}, June 2018.

\bibitem{PWC-net}
Deqing Sun, Xiaodong Yang, Ming-Yu Liu, and Jan Kautz.
\newblock Pwc-net: Cnns for optical flow using pyramid, warping, and cost
  volume.
\newblock {\em 2018 IEEE/CVF Conference on Computer Vision and Pattern
  Recognition}, Jun 2018.

\bibitem{video_rec_2}
Shuyang Sun, Zhanghui Kuang, Lu Sheng, Wanli Ouyang, and Wei Zhang.
\newblock Optical flow guided feature: A fast and robust motion representation
  for video action recognition.
\newblock In {\em Proceedings of the IEEE Conference on Computer Vision and
  Pattern Recognition (CVPR)}, June 2018.

\bibitem{R11}
Yao Tang, Lin Zhao, Shanshan Zhang, Chen Gong, Guangyu Li, and Jian Yang.
\newblock Integrating prediction and reconstruction for anomaly detection.
\newblock {\em Pattern Recognition Letters}, 129:123 -- 130, 2020.

\bibitem{M2}
F. Tung, J. Zelek, and David~A Clausi.
\newblock Goal-based trajectory analysis for unusual behaviour detection in
  intelligent surveillance.
\newblock {\em Image Vis. Comput.}, 29:230--240, 2011.

\bibitem{M1}
S. Wu, B.~E. Moore, and M. Shah.
\newblock Chaotic invariants of lagrangian particle trajectories for anomaly
  detection in crowded scenes.
\newblock {\em 2010 IEEE Computer Society Conference on Computer Vision and
  Pattern Recognition}, pages 2054--2060, 2010.

\bibitem{FramePredOpt3}
Yue Wu, Rongrong Gao, Jaesik Park, and Qifeng Chen.
\newblock Future video synthesis with object motion prediction.
\newblock {\em 2020 IEEE/CVF Conference on Computer Vision and Pattern
  Recognition (CVPR)}, pages 5538--5547, 2020.

\bibitem{M11}
D. Xu, E. Ricci, Yan Yan, Jingkuan Song, and N. Sebe.
\newblock Learning deep representations of appearance and motion for anomalous
  event detection.
\newblock {\em ArXiv}, abs/1510.01553, 2015.

\bibitem{R4}
D. Xu, Yan Yan, E. Ricci, and N. Sebe.
\newblock Detecting anomalous events in videos by learning deep representations
  of appearance and motion.
\newblock {\em Comput. Vis. Image Underst.}, 156:117--127, 2017.

\bibitem{R13}
Muchao Ye, Xiaojiang Peng, Weihao Gan, Wei Wu, and Yu Qiao.
\newblock Anopcn: Video anomaly detection via deep predictive coding network.
\newblock In {\em Proceedings of the 27th ACM International Conference on
  Multimedia}, MM '19, page 1805–1813, New York, NY, USA, 2019. Association
  for Computing Machinery.

\bibitem{object_centric_5}
Guang Yu, Siqi Wang, Zhiping Cai, En Zhu, Chuanfu Xu, Jianping Yin, and Marius
  Kloft.
\newblock Cloze test helps: Effective video anomaly detection via learning to
  complete video events.
\newblock {\em Proceedings of the 28th ACM International Conference on
  Multimedia}, Oct 2020.

\bibitem{AddCloseset_detail}
Guang Yu, Siqi Wang, Zhiping Cai, En Zhu, Chuanfu Xu, Jianping Yin, and Marius
  Kloft.
\newblock Cloze test helps: Effective video anomaly detection via learning to
  complete video events.
\newblock In {\em Proceedings of the 28th ACM International Conference on
  Multimedia}, MM '20, page 583–591, New York, NY, USA, 2020. Association for
  Computing Machinery.

\bibitem{M3}
Dong Zhang, D. Gatica-Perez, S. Bengio, and I. McCowan.
\newblock Semi-supervised adapted hmms for unusual event detection.
\newblock {\em 2005 IEEE Computer Society Conference on Computer Vision and
  Pattern Recognition (CVPR'05)}, 1:611--618 vol. 1, 2005.

\bibitem{M9}
B. Zhao, Li Fei-Fei, and E. Xing.
\newblock Online detection of unusual events in videos via dynamic sparse
  coding.
\newblock {\em CVPR 2011}, pages 3313--3320, 2011.

\bibitem{FPN_2}
Yiru Zhao, Bing Deng, Chen Shen, Yao Liu, Hongtao Lu, and Xian-Sheng Hua.
\newblock Spatio-temporal autoencoder for video anomaly detection.
\newblock In {\em Proceedings of the 25th ACM International Conference on
  Multimedia}, MM '17, page 1933–1941, New York, NY, USA, 2017. Association
  for Computing Machinery.

\bibitem{label_opt_2}
Jia-Xing Zhong, Nannan Li, Weijie Kong, Shan Liu, Thomas~H. Li, and Ge Li.
\newblock Graph convolutional label noise cleaner: Train a plug-and-play action
  classifier for anomaly detection.
\newblock In {\em Proceedings of the IEEE/CVF Conference on Computer Vision and
  Pattern Recognition (CVPR)}, June 2019.

\bibitem{R14}
J.~T. {Zhou}, J. {Du}, H. {Zhu}, X. {Peng}, Y. {Liu}, and R.~S.~M. {Goh}.
\newblock Anomalynet: An anomaly detection network for video surveillance.
\newblock {\em IEEE Transactions on Information Forensics and Security},
  14(10):2537--2550, 2019.

\bibitem{image_trans_1}
Jun-Yan Zhu, Richard Zhang, Deepak Pathak, Trevor Darrell, Alexei~A Efros,
  Oliver Wang, and Eli Shechtman.
\newblock Toward multimodal image-to-image translation.
\newblock In I. Guyon, U.~V. Luxburg, S. Bengio, H. Wallach, R. Fergus, S.
  Vishwanathan, and R. Garnett, editors, {\em Advances in Neural Information
  Processing Systems 30}, pages 465--476. Curran Associates, Inc., 2017.

\end{thebibliography}
}

\end{document}